\documentclass[letterpaper, 10 pt, conference]{ieeeconf}  

\IEEEoverridecommandlockouts                              

\usepackage{amsmath} 
\usepackage{amssymb}  
\usepackage{graphicx}
\graphicspath{{figures/}}
\usepackage{booktabs}
\usepackage[table]{xcolor}
\usepackage{subfigure}
\usepackage{stfloats}

\newcommand{\upcite}[1]{\textsuperscript{\cite{#1}}}

\title{\LARGE \bf
fwxRoute-Constrained Robust Fusion Estimation for MEMS/GNSS Integrated Navigation of Unmanned Ground Vehicles in GNSS Degraded Environments
}

\author{
Jingzhi Cui$^{1}$,
Chao Zhang$^{2}$,
Yuliang Mao$^{2}$,
Shaolin L{\"u}$^{1}$,
Dongmei Li$^{*1}$,
Huan Che$^{*2}$,
and Rong Zhang$^{*1}$%
\thanks{$^{1}$Jingzhi Cui, Shaolin L{\"u}, Dongmei Li, and Rong Zhang are with
State Key Laboratory of Precision Space-time Information Sensing Technology, Tsinghua University, Beijing, China.}
\thanks{$^{2}$Chao Zhang, Yuliang Mao, and Huan Che are with Xiaomi Inc., Beijing, China.}
\thanks{Corresponding authors: Dongmei Li, Huan Che, and Rong Zhang.}
}

\begin{document}

\maketitle
\thispagestyle{empty}
\pagestyle{empty}

\begin{abstract}

        To address cumulative localization drift of unmanned ground vehicles in structured road environments under 
        severe Global Navigation Satellite System signal occlusion, this paper proposes a robust route-constrained 
        state estimation method. During periods without satellite signals, the proposed method establishes the 
        correspondence between the historical dead reckoning trajectory and local segments of the mission route extracted
         from a high-definition map, and estimates a route-referenced position via a two-dimensional rigid transformation. 
         The estimated position is then formulated as a pseudo-position observation and incorporated into an Extended Kalman Filter update. 
         In this way, route constraints at the road level can be continuously injected into a unified state estimation framework, thereby
          suppressing position deviation relative to the mission route while indirectly improving azimuth estimation. To enhance practical
           applicability, engineering strategies, such as trigger control, matching quality validation, route offset compensation, and single update 
           correction limiting,  are further introduced. Experiments in three representative scenarios, including a long 
           tunnel, a multi-segment tunnel, and a curved tunnel, show that the proposed method effectively suppresses error accumulation
            during satellite outages, reduces the risk of large maximum deviation, and improves localization continuity and road-level 
            usability.

\end{abstract}

\section{INTRODUCTION}

In structured road environments, Unmanned Ground Vehicles (UGVs) usually rely on the fusion of the Global Navigation Satellite System (GNSS), the Inertial Measurement Unit (IMU), and wheel encoder measurements for global localization\upcite{wu2025wheelgins}. However, GNSS signals are often degraded or completely lost in environments such as tunnels, roads under viaducts, or areas with dense tree cover, leading to localization drift and azimuth errors due to reliance on dead reckoning. Maintaining localization continuity and suppressing drift in GNSS-degraded environments remains a practical challenge for UGVs\upcite{zhang2025faultygnss}.

Existing methods for handling GNSS degradation can generally be divided into two categories. One category improves state observability by introducing additional sensing sources, such as vision, Light Detection And Ranging Simultaneous Localization and Mapping (LiDAR-SLAM), Ultra-WideBand (UWB) localization, or vision and High-Definition (HD) map assisted localization. These methods enhance navigation robustness when GNSS signals are unavailable\upcite{zhang2024gnssfgo}. For example, LiDAR-SLAM can be tightly integrated with GNSS, IMU, and wheel odometry (ODO) to improve localization performance in tunnel environments\upcite{chang2020}. Vision aided localization has also shown effectiveness 
in challenging tunnel scenarios\upcite{ng2021shadow}. In addition, vision aided GNSS positioning has been investigated for autonomous systems in degraded urban environments\upcite{wen2023vision}. More recently, LiDAR-based fusion methods have continued to improve localization robustness under GNSS degradation\upcite{liu2025lidar}. UWB localization has also been considered as an alternative source of supplementary constraints\upcite{nguyen2025uloc}.

The second category exploits environmental priors, such as road geometry, lane centerlines, or mission routes, to impose map based constraints on localization results\upcite{ng2021mapping}. This method is attractive from an engineering perspective, particularly when the mission route is already available from an upstream navigation system. Early map-matching techniques demonstrated that route constraints could effectively improve localization consistency under GNSS challenged conditions\upcite{hunter2014}. More recently, GNSS/IMU integration and factor graph optimization have been used to improve robustness in GNSS-degraded environments\upcite{kassas2020}.

For UGVs in GNSS-degraded environments, the mission route often provides a useful prior even when GNSS signals are unavailable. The system can still retain a short-term historical dead reckoning trajectory based on IMU and wheel encoder measurements\upcite{wen2021fgo}. If this trajectory can be matched to a segment of the mission route, a pseudo-position observation can be constructed, which helps in maintaining localization continuity\upcite{wen2022gnc}. This approach does not require additional sensors and can continuously inject route constraints into the existing localization system\upcite{li2013}.

Motivated by these observations, this paper proposes a robust route-constrained state estimation method for GNSS-degraded environments. The method constructs a pseudo-position observation by matching a short-term historical dead reckoning trajectory with a local segment of the mission route from the HD map. This pseudo-position is incorporated into the Extended Kalman Filter (EKF) update, improving both position and azimuth estimation\upcite{zhang2022lio}. Furthermore, route offset compensation and quality control mechanisms are introduced to improve the robustness of the method in long tunnels and multi-segment occlusions\upcite{zhong2024trajectory}.

\begin{figure*}[t]
    \centering
    \includegraphics[width=0.88\textwidth]{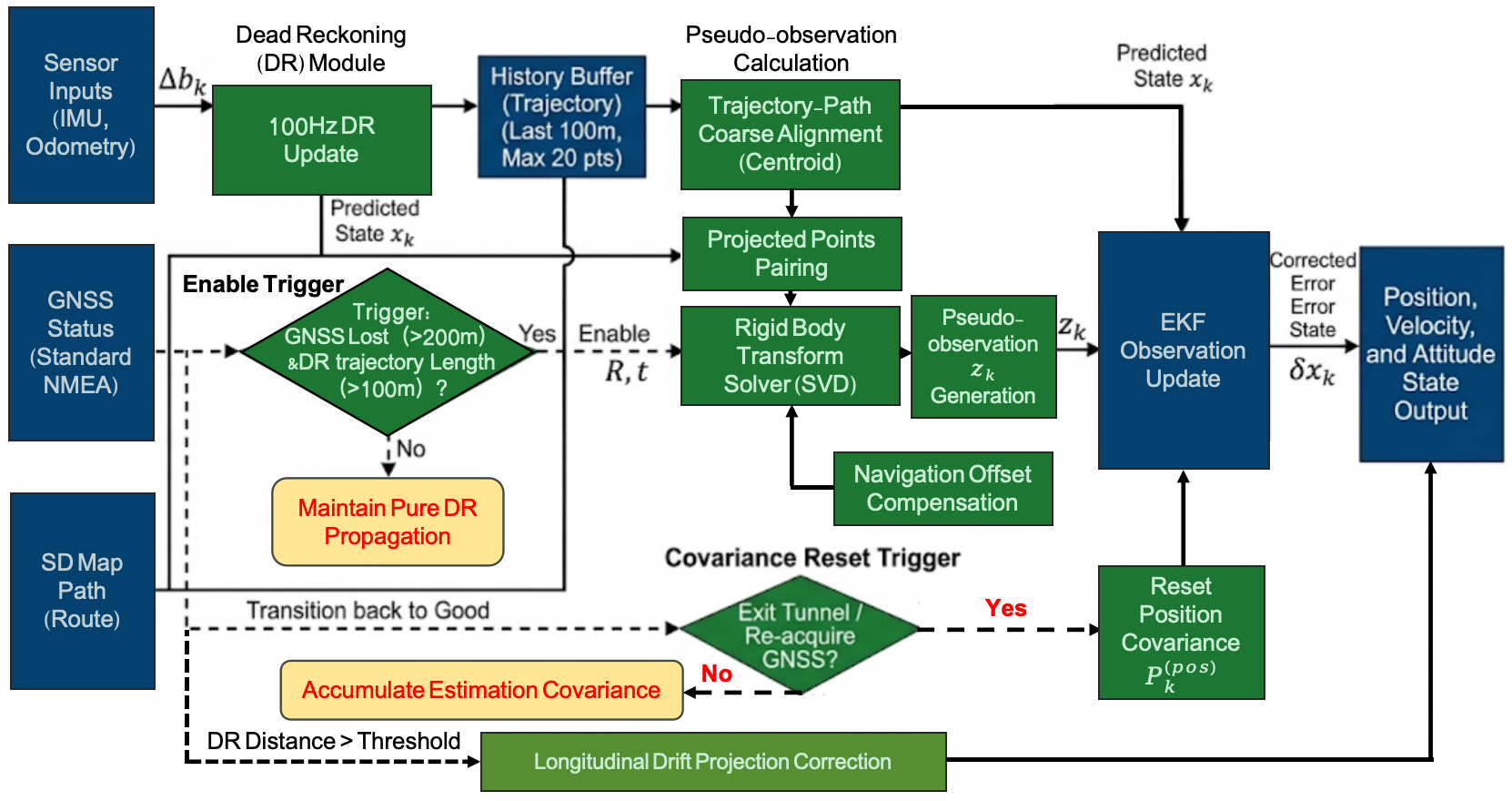}
    \caption{Overview of the proposed route-constrained state estimation framework.}
    \label{fig:framework}
    \vspace{-4pt}
\end{figure*}

The main contributions of this paper are summarized as follows:

(1) A robust route-constrained state estimation method is proposed for GNSS-unavailable periods. The method constructs a pseudo-position observation from the historical trajectory and injects it into the EKF update.

(2) A route offset compensation and quality control mechanism is introduced, ensuring the practical applicability of the method in real world scenarios with long GNSS outages and challenging environmental conditions.

\section{Pseudo-Position Observation Method Based on Trajectory Path Matching}

Under severe GNSS occlusion, UGVs cannot continuously obtain stable absolute position constraints. However, the system can still exploit two types of information: 
one is the short-term historical dead reckoning trajectory propagated by the IMU, wheel odometry, and related sensors; the other is the mission navigation path provided 
by the SD map. Based on these two sources of information, this paper adopts a pseudo-position observation method based on local trajectory-path matching. 
Rather than directly projecting the current position onto the mission navigation path for geometric resetting, the proposed method estimates a path reference
 position from the shape consistency between the short-term historical dead reckoning trajectory and the corresponding local segment of the mission navigation path, 
 and then constructs it as a pseudo-position observation that can be incorporated into the EKF update. By doing so, road-level path constraints are introduced
  within a unified state estimation framework.

It should be noted that, although the original system usually represents UGV positions and navigation paths in geographic coordinates, 
the method proposed in this paper is modeled in a local planar coordinate system to facilitate trajectory-path local matching, 
pseudo-position observation construction, and filtering update. Accordingly, the route-matching constraint in this paper
 is formulated only in the horizontal plane, and altitude is not explicitly constrained by the navigation path prior.

\begin{figure*}[t]
    \centering
    \includegraphics[width=0.88\linewidth]{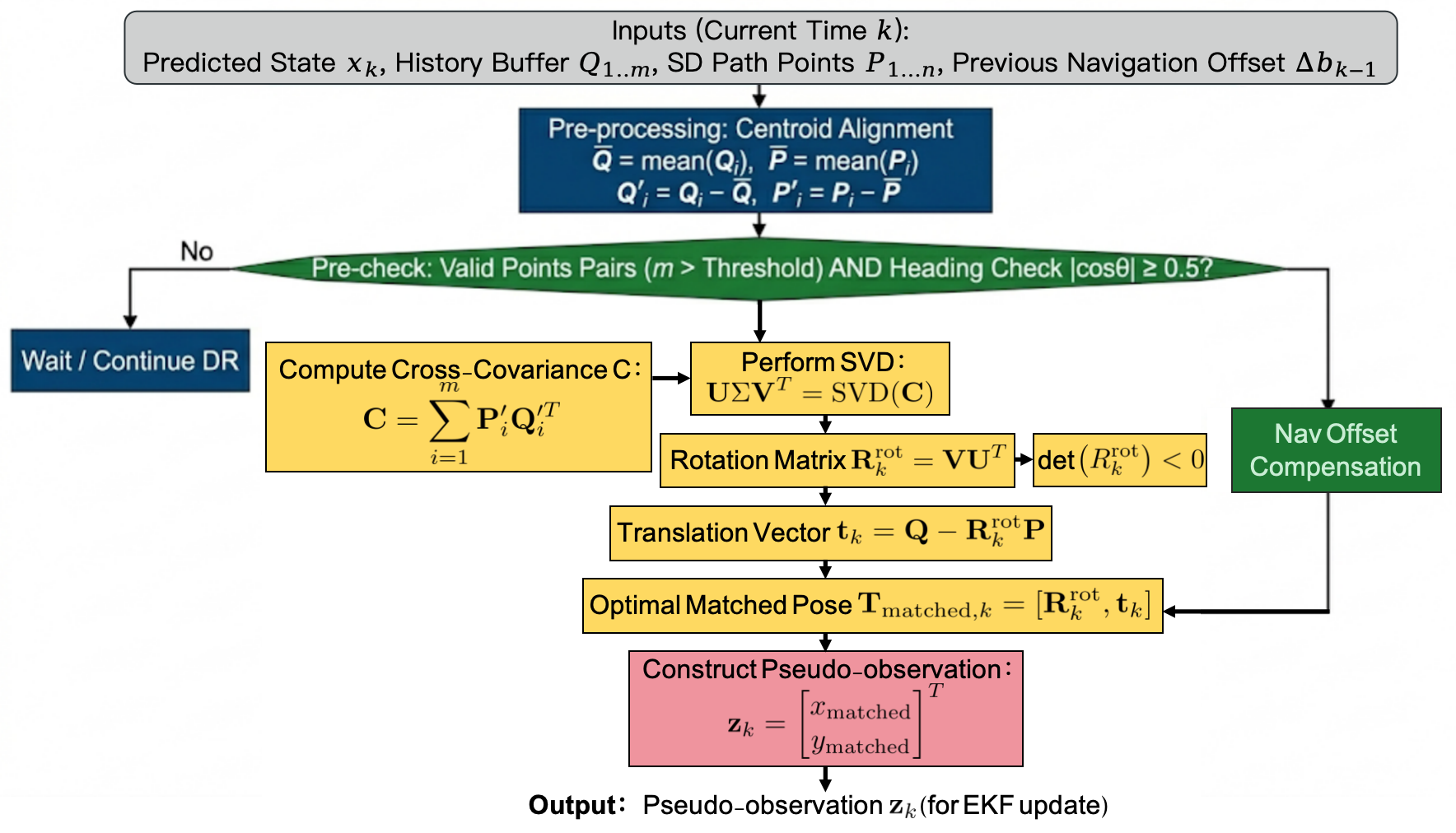}
    \caption{Flowchart of pseudo-position observation construction based on local trajectory-path matching.}
    \label{fig:matching_flow}
    \vspace{-4pt}
\end{figure*}

\subsection{Local Trajectory Path Matching and Path Reference Position Construction}

To avoid mismatching when GNSS fluctuates briefly, the historical trajectory is too short, or the local geometric information is insufficient,
 path constraints are activated only when certain conditions are satisfied. 
 Specifically, a low frequency pseudo-position update is performed only after GNSS has been continuously unavailable for a period of time and a sufficiently
  long historical trajectory has been accumulated during the GNSS-unavailable interval.

At each valid trigger instant, a segment of historical dead reckoning trajectory before the current time is extracted and denoted as
\begin{equation}
\mathcal{T}_k=\left\{\mathbf{p}_{k-L+1},\mathbf{p}_{k-L+2},\ldots,\mathbf{p}_k\right\}
\end{equation}
where $\mathbf{p}_k=[p_{E,k},\,p_{N,k}]^T$ is the position estimate at time $k$ in the local East--North plane, and $L$ is the historical trajectory window length. To reduce the influence of nonuniform sampling and speed fluctuations, the historical trajectory is uniformly discretized at a fixed spatial interval. Meanwhile, a local segment of the mission navigation path around the current position is extracted as
\begin{equation}
\mathcal{N}_k=\left\{\mathbf{q}_1,\mathbf{q}_2,\ldots,\mathbf{q}_M\right\}
\end{equation}
where $\mathbf{q}_i \in \mathbb{R}^2$ denotes the $i$th discretized point on the local mission navigation path segment, and $M$ is the number of discretized points in that local path segment. This path segment is discretized in the same manner as the historical trajectory.
Point correspondences are then established according to the cumulative arc-length order:
\begin{equation}
\left\{(\mathbf{s}_i,\mathbf{t}_i)\right\}_{i=1}^{N}
\end{equation}
where $\mathbf{s}_i\in\mathbb{R}^2$
 denotes a discretized point on the historical trajectory and $\mathbf{t}_i\in\mathbb{R}^2$ denotes the corresponding point on the local mission route segment.

In Fig.~\ref{fig:matching_flow}, $m$ denotes the number of valid point pairs after spatial resampling and correspondence construction. The matching quality is evaluated by the number of valid point pairs, terminal heading consistency, and rigid-registration residual. A pseudo-position observation is constructed only when these criteria are satisfied; otherwise, the route-constrained update is skipped.

To reduce the risk of mismatching to adjacent parallel lanes or opposite direction route segments, an azimuth-consistency check is first performed before rigid alignment. 
The resampled dead reckoning trajectory segment and its projected route correspondences within the local window can be denoted by
\begin{align}
\mathcal{S}_k &= \{\mathbf{s}_i\}_{i=1}^{N} \\
\mathcal{T}_k &= \{\mathbf{t}_i\}_{i=1}^{N}
\end{align}
where $\mathbf{s}_i,\mathbf{t}_i \in \mathbb{R}^2$ are expressed in the local plane. Since both sets are resampled with a fixed spatial interval, the tangent direction near the trajectory end can be approximated by the line segment formed by the last two points. The normalized terminal direction vectors are defined as
\begin{equation}
\hat{\mathbf{d}}_{S,k}
=
\frac{\mathbf{s}_{N}-\mathbf{s}_{N-1}}
{\left\|\mathbf{s}_{N}-\mathbf{s}_{N-1}\right\|}
\qquad
\hat{\mathbf{d}}_{T,k}
=
\frac{\mathbf{t}_{N}-\mathbf{t}_{N-1}}
{\left\|\mathbf{t}_{N}-\mathbf{t}_{N-1}\right\|}
\label{eq:heading_vec}
\end{equation}

The azimuth consistency is then evaluated by
\begin{equation}
\cos\theta_k
=
\hat{\mathbf{d}}_{S,k}^{\top}\hat{\mathbf{d}}_{T,k}
\label{eq:heading_consistency}
\end{equation}

Only candidate matches satisfying
$|\cos\theta_k| \ge \gamma_h$
are retained, where $\gamma_h$ is a predetermined threshold.
 This screening step helps reject mismatches caused by reverse-direction lanes, neighboring parallel roads, 
 or locally inconsistent route topologies.

After passing the azimuth-consistency test, the route-constrained alignment is formulated as a two-dimensional rigid
 registration problem. The optimal rotation matrix $\mathbf{R}^{\mathrm{rot}}_k \in SO(2)$ and translation vector 
 $\mathbf{t}_k \in \mathbb{R}^2$, subject to $\mathbf{R}_k^{\mathrm{rot}\top}\mathbf{R}^{\mathrm{rot}}_k=\mathbf{I}$ and $\det(\mathbf{R}^{\mathrm{rot}}_k)=1$, are obtained by solving
\begin{equation}
(\mathbf{R}^{\mathrm{rot}}_k,\mathbf{t}_k)
=
\arg\min_{\mathbf{R}^{\mathrm{rot}}_k,\mathbf{t}_k}
\sum_{i=1}^{N}
\left\|
\mathbf{R}^{\mathrm{rot}}_k \mathbf{s}_i + \mathbf{t}_k - \mathbf{t}_i
\right\|_2^2
\label{eq:rigid_registration}
\end{equation}

To compute the closed-form solution, the centroids of the two point sets are first defined as
\begin{equation}
\bar{\mathbf{s}}_k=
\frac{1}{N}\sum_{i=1}^{N}\mathbf{s}_i,
\qquad
\bar{\mathbf{t}}_k=
\frac{1}{N}\sum_{i=1}^{N}\mathbf{t}_i
\label{eq:centroids}
\end{equation}

To suppress accumulated longitudinal drift during prolonged GNSS outages, a projection-based correction is applied before outputting the position estimate.

The cross covariance matrix is then constructed as
\begin{equation}
\mathbf{C}_k=
\sum_{i=1}^{N}
(\mathbf{s}_i-\bar{\mathbf{s}}_k)
(\mathbf{t}_i-\bar{\mathbf{t}}_k)^{T}
\label{eq:cross_covariance}
\end{equation}

By applying singular value decomposition (SVD),
\begin{equation}
\mathbf{C}_k=\mathbf{U}_k \boldsymbol{\Sigma}_k \mathbf{V}_k^{T}
\label{eq:svd}
\end{equation}
the optimal rotation and translation can be written as
\begin{equation}
\mathbf{R}^{\mathrm{rot}}_k
=
\mathbf{V}_k
\begin{bmatrix}
1 & 0\\
0 & \det(\mathbf{V}_k \mathbf{U}_k^{T})
\end{bmatrix}
\mathbf{U}_k^{T}
\label{eq:rotation_solution}
\end{equation}

\begin{equation}
\mathbf{t}_k
=
\bar{\mathbf{t}}_k
-
\mathbf{R}^{\mathrm{rot}}_k \bar{\mathbf{s}}_k
\label{eq:translation_solution}
\end{equation}
        
\begin{figure*}[t]
    \centering
    \subfigure[Initial Azimuth Error and Translation Error]{
        \includegraphics[width=0.3\textwidth]{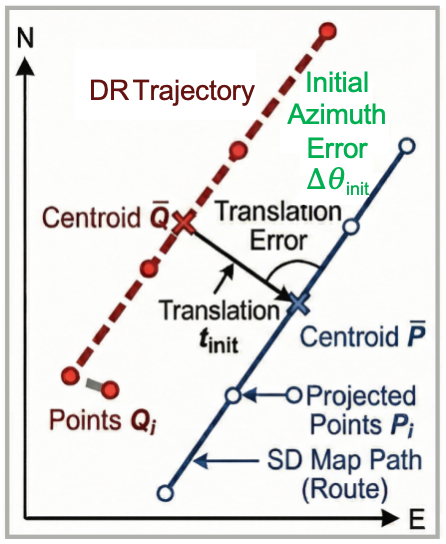}
        \label{fig:part1}
    }
    \subfigure[Centroid Alignment]{
        \includegraphics[width=0.3\textwidth]{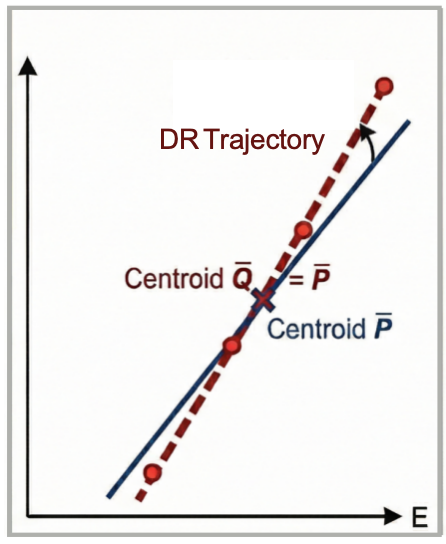}
        \label{fig:part2}
    }
    \subfigure[Pseudo-Observation Construction]{
        \includegraphics[width=0.3\textwidth]{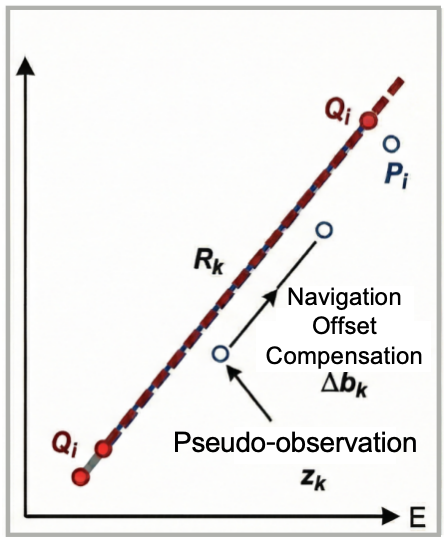}
        \label{fig:part3}
    }
    \caption{Geometric illustration of trajectory-path alignment, route offset compensation, and pseudo-position observation construction.}
    \label{fig:geometric_illustration}
    \vspace{-4pt}
\end{figure*}

Let $\mathbf{p}_k^{\mathrm{dr}}$ denote the current dead reckoning position estimate in the ENU frame. Since $(\mathbf{R}^{\mathrm{rot}}_k,\mathbf{t}_k)$ is estimated from the historical trajectory window and its corresponding local route segment, the route-referenced position at the current epoch is constructed as
\begin{equation}
\hat{\mathbf{p}}_k^{\mathrm{ref}}
=
\mathbf{R}^{\mathrm{rot}}_k \mathbf{p}_k^{\mathrm{dr}} + \mathbf{t}_k
\label{eq:route_referenced_position}
\end{equation}

This quantity is not treated as an absolute ground truth position; rather, it is a locally route-referenced position inferred
 from the geometric consistency between the recent dead reckoning trajectory and the mission route segment.

\subsection{State Space Model and Pseudo-Position Observation Update}

To incorporate the above path reference position into the filter in a unified measurement update form, this paper formulates
 the method within the error state EKF framework. Since the original system already has a prediction mechanism based 
 on inertial sensors and wheel odometry, this paper does not modify its propagation model.

During GNSS-unavailable periods, the original system degrades to dead reckoning based mainly on
 inertial and wheel encoder measurements. Its propagation model can be written as
\begin{equation}
\mathbf{x}_k^- = f(\mathbf{x}_{k-1}^+,\mathbf{u}_k)
\end{equation}

Without additional route-constrained observations, the filter mainly relies on prediction during outage intervals, i.e.,
\begin{equation}
\mathbf{x}_k^+ \approx \mathbf{x}_k^-
\end{equation}

Consequently, position drift builds up incrementally due to cumulative azimuth and odometry errors.
 The proposed method keeps this propagation model unchanged and introduces an additional pseudo-position observation only 
 in the update phase.

After GNSS re-acquisition, a covariance reset is applied to avoid abrupt position jumps during the transition back to standard fusion.

The 15-dimensional error state vector is defined as
\begin{equation}
\delta \mathbf{x}=
\begin{bmatrix}
\delta \boldsymbol{\phi} \\
\delta \mathbf{v} \\
\delta \mathbf{p} \\
\delta \mathbf{b}_{g} \\
\delta \mathbf{b}_{a}
\end{bmatrix}
\in \mathbb{R}^{15}
\label{eq:error_state}
\end{equation}
where $\delta \boldsymbol{\phi}$ denotes the attitude error vector, 
$\delta \mathbf{v}$ denotes the velocity error vector, 
$\delta \mathbf{p}=[\delta p_E,\delta p_N,\delta p_U]^T$ denotes the position error vector, 
and $\delta \mathbf{b}_{g}$ and $\delta \mathbf{b}_{a}$ denote the gyroscope and accelerometer bias errors, respectively.
It should be noted that, although the error state model is maintained in a full 3D form, the route-matching constraint considered in this paper is inherently planar.
 Since the task navigation path provides a road-level reference mainly in the horizontal directions,
  the pseudo-position observation constructed 
 in this work constrains only the east and north position components, while the vertical position component is
  not directly corrected in this update.

Within the error state framework, the continuous-time propagation model is written as
\begin{equation}
\delta \dot{\mathbf{x}}
=
\mathbf{F} \delta \mathbf{x}
+
\mathbf{w}
\label{eq:continuous_model}
\end{equation}
where $\mathbf{F}$ is the state transition matrix, 
and $\mathbf{w}\sim N(0,\mathbf{Q})$ is the 
process noise.
 According to inertial navigation error propagation, $\mathbf{F}_k$ can be written as
\begin{equation}
\mathbf{F}=
\begin{bmatrix}
\mathbf{F}_{\phi\phi} & \mathbf{F}_{\phi v} & \mathbf{F}_{\phi p} & -\mathbf{C}_b^n & \mathbf{0} \\
\mathbf{F}_{v\phi} & \mathbf{F}_{vv} & \mathbf{F}_{vp} & \mathbf{0} & \mathbf{C}_b^n \\
\mathbf{0} & \mathbf{F}_{pv} & \mathbf{F}_{pp} & \mathbf{0} & \mathbf{0} \\
\mathbf{0} & \mathbf{0} & \mathbf{0} & -\frac{1}{\tau_g}\mathbf{I}_3 & \mathbf{0} \\
\mathbf{0} & \mathbf{0} & \mathbf{0} & \mathbf{0} & -\frac{1}{\tau_a}\mathbf{I}_3
\end{bmatrix}
\label{eq:Fk}
\end{equation}
where $\mathbf{0}$ denotes a zero matrix of compatible dimensions, $\mathbf{C}_b^n$ is the direction cosine matrix from the body 
frame to the navigation frame,
 and $\tau_g$ and $\tau_a$ are the correlation time constants of the first-order Markov processes for gyroscope and 
 accelerometer biases, respectively. 
This model indicates that, when GNSS is unavailable, attitude, velocity, and position errors are mutually coupled and gradually accumulate during propagation. After discretization, the error-state propagation model from time step $k-1$ to time step $k$ can be written as
\begin{equation}
\delta \mathbf{x}_{k|k-1}
=
\boldsymbol{\Phi}_{k-1}\,\delta \mathbf{x}_{k-1|k-1},
\label{eq:discrete_model}
\end{equation}
where $\delta \mathbf{x}_{k|k-1}$ denotes the predicted error-state vector at time step $k$ based on the information available up to time step $k-1$, $\delta \mathbf{x}_{k-1|k-1}$ denotes the updated error-state vector at time step $k-1$, and $\boldsymbol{\Phi}_{k-1}$ is the discrete-time state transition matrix over the interval from time step $k-1$ to time step $k$. Under first-order discretization, $\boldsymbol{\Phi}_{k-1}$ can be approximated as
\begin{equation}
\boldsymbol{\Phi}_{k-1}
\approx
\mathbf{I}+\mathbf{F}_{k-1}\,\Delta t,
\label{eq:Phi_discrete}
\end{equation}
where $\mathbf{F}_{k-1}$ is the continuous-time error-state matrix evaluated at time step $k-1$, and $\Delta t$ is the sampling interval.

\begin{figure*}[t]
    \centering
    \subfigure[Tunnel and Surrounding Area from Google Maps.]{
        \includegraphics[width=0.45\textwidth]{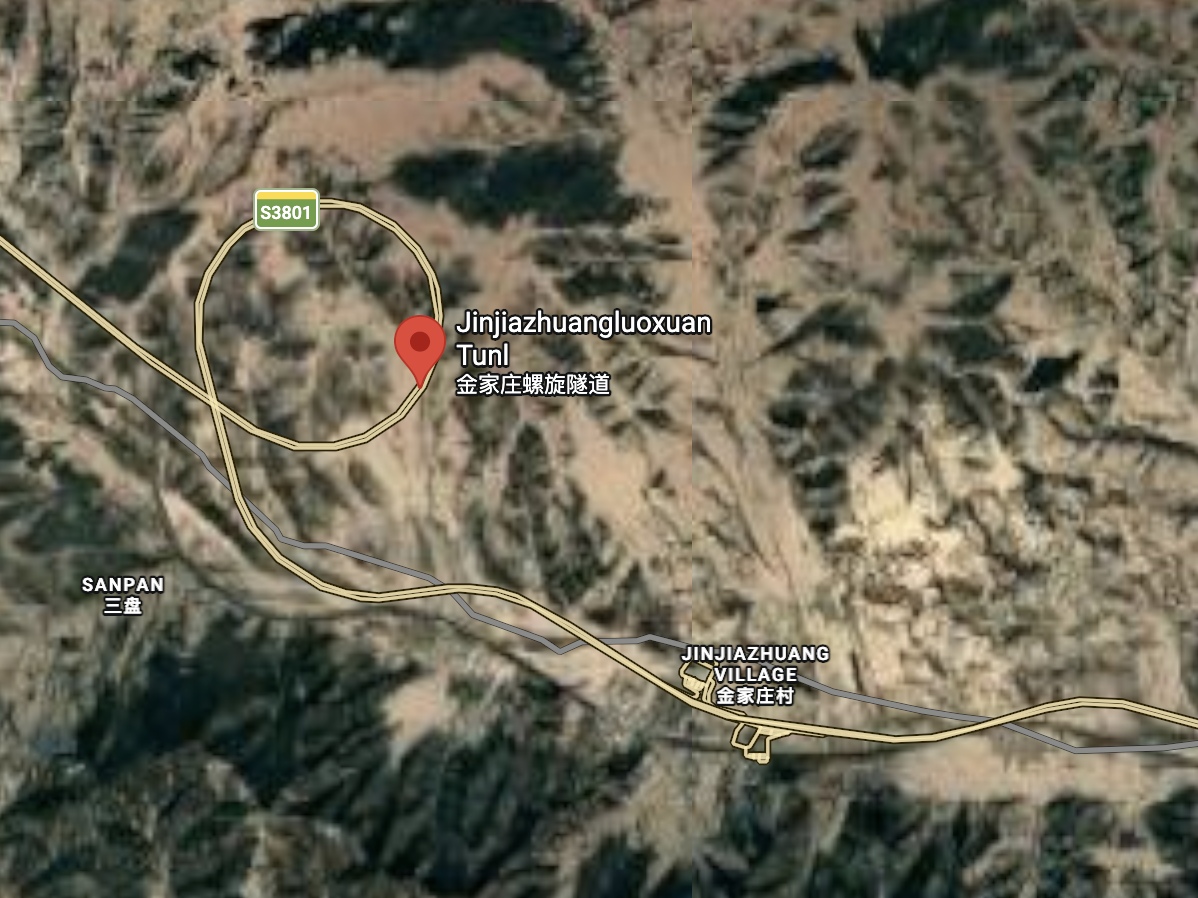}
        \label{fig:map_view}
    }
    \subfigure[Tunnel Entrance.]{
        \includegraphics[width=0.45\textwidth]{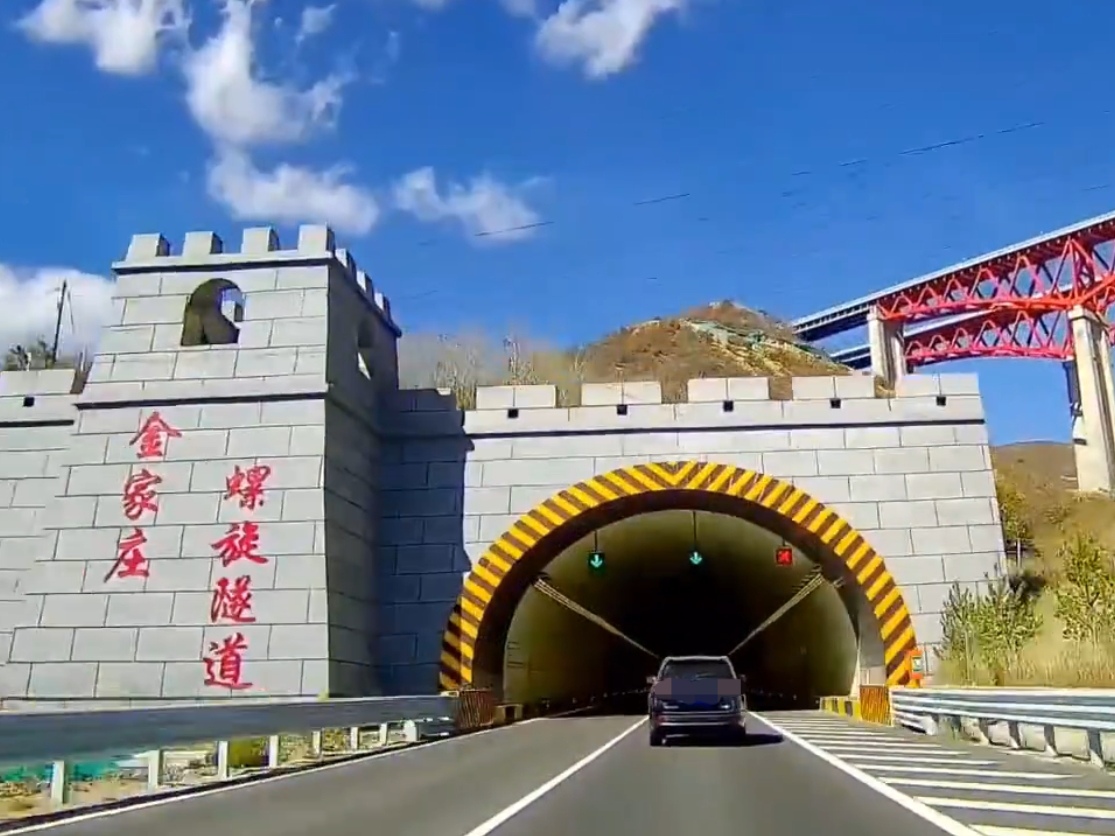}
        \label{fig:tunnel_entrance}
    }
    \caption{Geometric illustration of the tunnel scenario.}
    \label{fig:tunnel_scenario}
    \vspace{-4pt}
\end{figure*}

In the observation update, the proposed method does not directly use real GNSS measurements. 
Instead, the path reference generated by local trajectory-path matching is further transformed
 into a pseudo-position observation. Since the mission navigation path usually provides only a road-level reference,
  directly using $\hat{\mathbf p}_k^{\mathrm{ref}}$ as an accurate position measurement may force the state estimate unnaturally
   toward the path centerline. To alleviate this effect, a historical offset compensation term $\Delta \mathbf{b}_k$ is introduced,
    and the pseudo-position observation is constructed as
\begin{equation}
\mathbf{z}_k=\hat{\mathbf p}_k^{\mathrm{ref}}+\Delta \mathbf{b}_k
\label{eq:zk}
\end{equation}
where $\mathbf{z}_k$ denotes the pseudo-position observation at time step $k$, 
$\hat{\mathbf{p}}_k^{\mathrm{ref}}$ denotes the route-referenced position obtained from local trajectory-path matching at time step $k$, 
and $\Delta \mathbf{b}_k$ denotes the navigation offset compensation term at time step $k$.

Here, $\Delta \mathbf{b}_k$ is used to preserve, as much as possible, the UGV's stable lateral offset relative to the mission 
route,
 so that the path constraint acts as a weak road-level constraint rather than an absolute ground truth position constraint.

This is because the mission route used in this paper serves as a road-level planar reference rather than a full 3D 
geometric prior. Therefore, no direct altitude observation is introduced in the route-constrained update. 
When only planar position constraints are considered, the measurement model can be written as
\begin{equation}
\mathbf{z}_k=h(\delta \mathbf{x}_k)+\mathbf{v}_k
\label{eq:measurement_model}
\end{equation}
where $\mathbf{v}_k$ is the observation noise. Since the constructed pseudo-observation constrains only the planar
 position components, the observation function $h(\cdot)$ is defined as
\begin{equation}
h(\delta \mathbf{x}_k)=
\begin{bmatrix}
\delta p_E\\
\delta p_N
\end{bmatrix}
\label{eq:hx}
\end{equation}
Accordingly, the observation matrix is
\begin{equation}
\mathbf{H}_k=
\left[
\mathbf{0}_{2\times 3}\;
\mathbf{0}_{2\times 3}\;
\mathbf{I}_{2\times 2}\;
\mathbf{0}_{2\times 4}\;
\mathbf{0}_{2\times 3}
\right]
\label{eq:Hk}
\end{equation}

Here, $\mathbf{R}_k$ denotes the covariance of the pseudo-position observation and is conservatively assigned to avoid over-constraining the dead-reckoning propagation. Matching results with large residuals or weak route geometry are rejected before the EKF update.

The EKF update is then given by
\begin{equation}
    \mathbf{P}_{k|k-1}=\boldsymbol{\Phi}_{k-1} \mathbf{P}_{k-1|k-1}\boldsymbol{\Phi}^T_{k-1}+\mathbf{Q}_k
\end{equation}
\begin{equation}
\mathbf{K}_k=
\mathbf{P}_{k|k-1}\mathbf{H}_k^T
\left(
\mathbf{H}_k\mathbf{P}_{k|k-1}\mathbf{H}_k^T+\mathbf{R}_k
\right)^{-1}
\label{eq:Kk}
\end{equation}
\begin{equation}
\delta \mathbf{x}_{k|k}=
\delta \mathbf{x}_{k|k-1}
+\mathbf{K}_k
\left(
\mathbf{z}_k-\mathbf{H}_k\delta \mathbf{x}_{k|k-1}
\right)
\label{eq:x_update}
\end{equation}
\begin{equation}
\mathbf{P}_{k|k}=
\left(
\mathbf{I}-\mathbf{K}_k\mathbf{H}_k
\right)\mathbf{P}_{k|k-1}
\label{eq:P_update}
\end{equation}

These equations show that the pseudo-position observation directly corrects the east and north position errors,
 while the remaining states, including attitude, velocity, and sensor biases, are adjusted indirectly through 
 state coupling in the covariance.

\subsection{Engineering Strategies and Quality Control}

To improve stability and engineering applicability in real road scenarios, several practical strategies are introduced during pseudo-position
 observation construction and update. First, the pseudo-position update frequency is limited through trigger control, so 
 that frequent corrections are avoided when short-term fluctuations occur or local information is insufficient. Second, direction-consistency
  checking and local registration residual constraints are used to reject clearly unreasonable matching results. Finally, an upper bound is imposed 
  on the magnitude of each single correction to avoid injecting abnormal pseudo-positions into the filter when the local geometry is insufficient, 
  the trajectory drift is too large, or the candidate path segment is unstable.

In summary, the proposed method does not directly project the current position onto the mission navigation path.
 Instead, it constructs a pseudo-position observation from the shape consistency between the short-term historical dead reckoning 
 trajectory and the local mission navigation path segment, and then suppresses the accumulated drift through EKF update. In this way, 
 the engineering stability of the original propagation model is preserved, while road-level path constraints are incorporated into the system
  in a unified state estimation form.

The proposed method may become less reliable in geometrically ambiguous scenarios, such as long straight roads, parallel tunnels, ramps, loops, and intersections, or when the mission route has meter-level bias relative to the actually driven lane. In such cases, unreliable updates are rejected based on valid point pairs, heading consistency, and registration residual, while robust route-candidate validation and adaptive confidence modeling will be investigated in future work.

\section{Experimental Evaluation}

To validate the effectiveness of the proposed method under severe GNSS degradation, experiments were conducted in 
three representative scenarios, namely a long tunnel, a multi-segment tunnel, and a curved tunnel. All three datasets
 were collected from the same real tunnel environment, i.e., the Jinjiazhuang extra-long spiral tunnel on the Jingli Expressway (Yanchong Expressway) near
  Chicheng County, Zhangjiakou, Hebei Province. The three files correspond to different local segments within the same tunnel.
   This data source ensures a consistent road-structure background and realistic GNSS challenged conditions across all scenarios,
    which is beneficial for a fair comparison of different localization methods.

\begin{figure}[!htbp]
    \centering
    \includegraphics[width=0.9\linewidth]{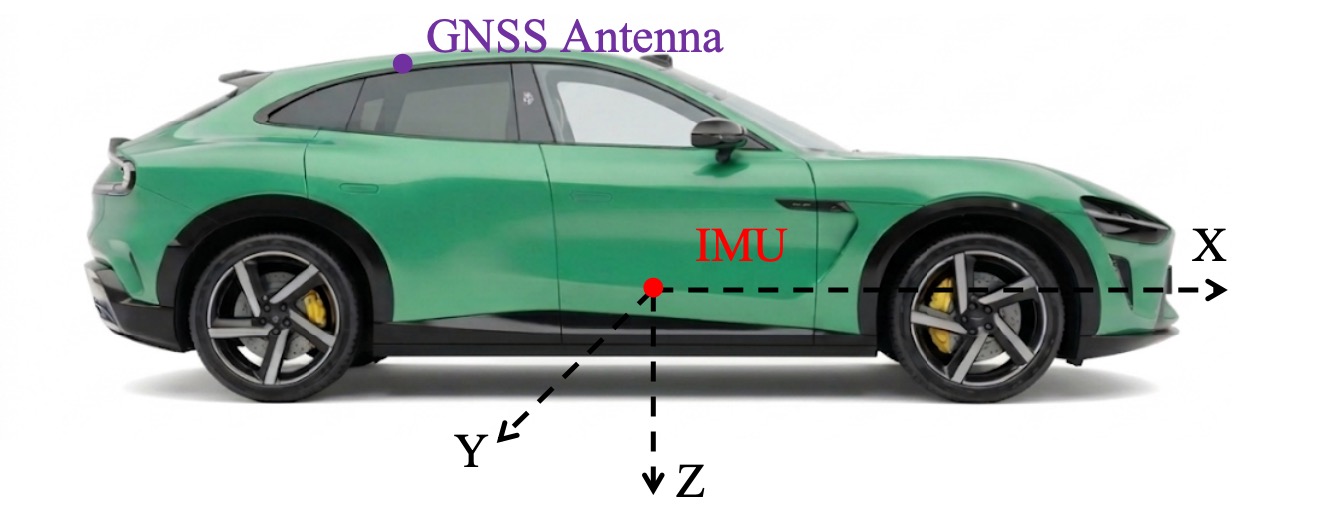}
    \caption{GNSS and IMU coordinate system for the vehicle platform.}
    \label{fig:vehicle_gnss_imu}
    \vspace{-4pt}
\end{figure}

The vehicle platform is supported by the IMU-BOX provided by Huace Navigation Inc. The vehicle image shown is a Xiaomi YU7.

Since the route used in this paper is derived from the mission route in the SD map, it serves as a road-level reference rather than absolute ground truth. Therefore, the reported position deviation should be interpreted as route-relative deviation, which is used to evaluate localization continuity and route-level usability during GNSS outages.

The compared methods include the original localization result (baseline) and the localization result with pseudo-position observations (proposed). The baseline refers to the degraded mode of the original GNSS/INS/ODO system during GNSS-unavailable intervals, where the solution mainly relies on inertial propagation and wheel-odometry-based dead reckoning. In contrast, the Proposed method incorporates the trajectory-to-route matching result as a pseudo-position observation into the EKF update.

\begin{figure}[!htbp]
    \centering
    \includegraphics[width=0.95\linewidth]{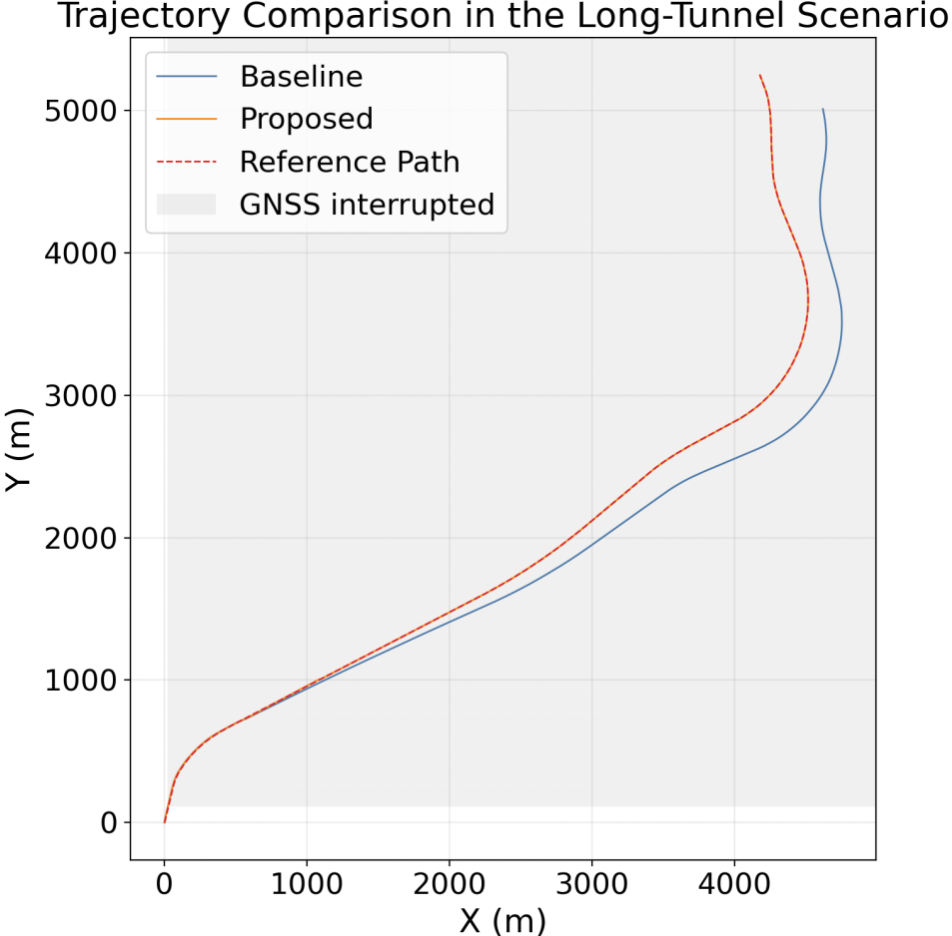}
    \caption{Trajectory comparison in the long-tunnel scenario.} 
    \label{fig:long_tunnel_trajectory}
    \vspace{-4pt}
\end{figure}

\begin{figure*}[t]
    \centering
    \includegraphics[width=0.96\textwidth]{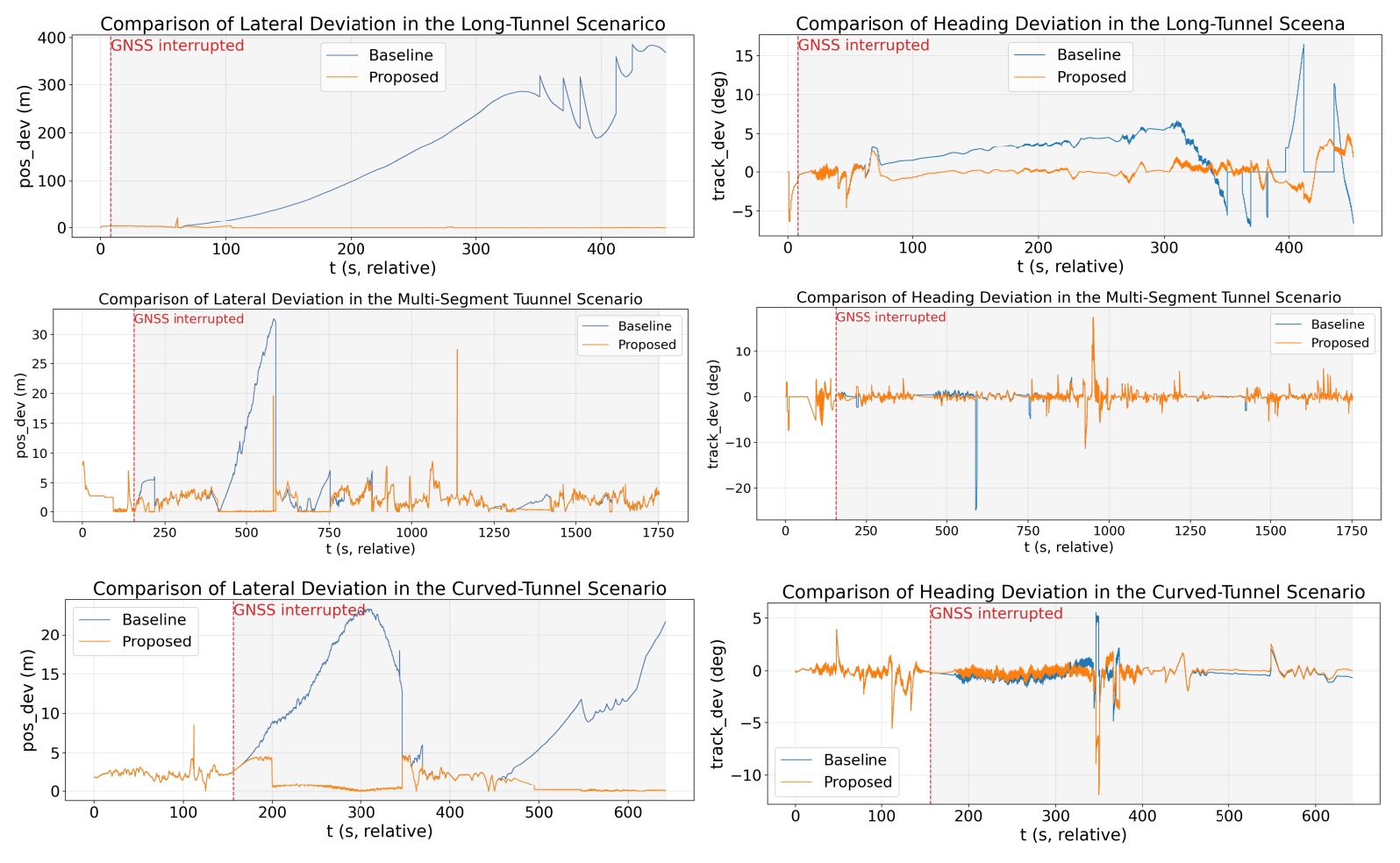}
    \caption{Comparison of position deviation and azimuth deviation between the baseline and the proposed method in three representative tunnel scenarios, including the long-tunnel, multi-segment tunnel, and curved-tunnel cases. The shaded region indicates the GNSS degraded interval.}
    \label{fig:deviation_comparison}
    \vspace{-4pt}
\end{figure*}

As shown in Fig.~\ref{fig:long_tunnel_trajectory}, in the long-tunnel scenario, the baseline gradually departs from the mission route after
 GNSS interruption and exhibits persistent accumulated drift. By contrast, the proposed method remains much closer to the reference
  path during the outage interval, indicating that the constructed pseudo-position observation can provide effective route constraints when 
  GNSS is unavailable. To further examine the behavior of the two methods in different scenarios, Fig.~\ref{fig:deviation_comparison} compares
   the time-series results of position  and azimuth deviations.

As shown in Fig.~\ref{fig:deviation_comparison}, the proposed method exhibits smaller position deviations and more stable azimuth variations in
 all three scenarios. In this figure, the shaded gray region denotes the GNSS degraded interval. These results suggest that the proposed method 
 can not only suppress position drift relative to the mission route, but also mitigate azimuth-error accumulation to some extent through 
 the state correlation in the filtering framework.
 
\begin{table*}[!b]
\centering
\caption{Quantitative comparison of route-relative position and heading deviations in three representative scenarios.}
\label{tab:scenario_comparison}
\renewcommand{\arraystretch}{1.2}
\setlength{\tabcolsep}{8pt}
\begin{tabular}{l l c c c c c}
\toprule
\textbf{Scenario} & \textbf{Method} & 
\textbf{Max Position Dev.} & 
\textbf{Mean Position Dev.} & 
\textbf{Position Dev. RMSE} & 
\textbf{Mean Heading Dev.} & 
\textbf{Heading Dev. RMSE} \\
\midrule
Long  & Baseline & 386.3 & 142.909 & 186.821 & 2.073  & 3.524 \\
\rowcolor{gray!20}
      & Proposed & 22.7  & 0.745   & 1.672   & 0.048  & 1.257 \\
\midrule
Curve & Baseline & 32.6  & 7.908   & 10.376  & -0.216 & 0.793 \\
\rowcolor{gray!20}
      & Proposed & 27.5  & 1.431   & 1.889   & -0.179 & 0.946 \\
\midrule
Multi & Baseline & 23.3  & 3.576   & 6.502   & 0.011  & 1.839 \\
\rowcolor{gray!20}
Segment    & Proposed & 8.5   & 1.755   & 2.351   & 0.018  & 1.332 \\
\bottomrule
\end{tabular}
\vspace{-4pt}
\end{table*}
In the long-tunnel scenario shown in Fig.~\ref{fig:deviation_comparison}, the position deviation of the baseline continues to grow after GNSS outage and gradually forms 
obvious accumulation, whereas the proposed method remains at a relatively low level throughout the whole interval,
 with significantly reduced fluctuation. The corresponding azimuth-deviation curves also show that 
 the Baseline exhibits a more obvious departure in the middle and later parts, while the proposed method
  stays overall closer to the zero-bias region. These results indicate that, under long-duration continuous GNSS blockage, 
  the proposed method can continuously introduce mission-route constraints and effectively suppress the accumulated divergence 
  of dead reckoning errors.

In the multi-segment tunnel scenario of Fig.~\ref{fig:deviation_comparison}, the position deviation of the baseline increases locally in multiple blocked intervals 
and forms relatively large peaks in some segments. The Proposed method maintains lower position deviation for most of the time, although the improvement
 is less pronounced than that in the long-tunnel scenario. The azimuth-deviation results show a similar trend: both methods fluctuate within a relatively
  small range for most of the time, but the proposed method recovers faster after abnormal fluctuations. This suggests that when GNSS blockage is intermittent, 
  some external constraints can still be periodically recovered between adjacent blocked segments, and the original solution itself does not exhibit persistent 
  divergence. Therefore, the main contribution of the proposed method in this scenario lies in further reducing local peak deviations.

In the curved-tunnel scenario shown in Fig.~\ref{fig:deviation_comparison}, the position deviation of the baseline gradually increases after GNSS 
outage and shows obvious accumulation in curved segments, whereas the proposed method consistently remains at a lower level with smaller overall fluctuations.
 The azimuth-deviation curves show that both methods fluctuate during some local intervals, but the proposed method stays closer to the zero-bias region for 
 most of the time. This demonstrates that even when the route geometry changes, the proposed method can still provide stable reference constraints for the
  current position through the matching relationship between the historical trajectory and the local task-route segment, thereby alleviating the lateral 
  offset caused by the coupling effect of azimuth errors.

As summarized in Table~\ref{tab:scenario_comparison}, the Proposed method achieves better quantitative results than the Baseline in all three scenarios. Specifically, the route-relative position-deviation metrics in Table~\ref{tab:scenario_comparison} are expressed in meters (m), whereas the heading-deviation metrics are expressed in degrees (deg). Overall,
 both the mean position deviation and the position RMSE are significantly reduced, indicating that the proposed method can more effectively suppress 
 the accumulated deviation relative to the mission route and reduce the fluctuation range of localization errors. Among the three scenarios, the improvement 
 is most significant in the long-tunnel case, suggesting that when GNSS remains unavailable for a long duration, the constructed pseudo-position observation
  can continuously provide effective route constraints and substantially enhance the stability of the position solution. By comparison, the improvements 
  in the curved-tunnel and multi-segment tunnel scenarios are relatively moderate, but a consistent improvement trend is still observed.

In terms of azimuth related metrics, the proposed method also shows better overall stability, although the improvement is more limited than that
 in the position related metrics. This indicates that the primary benefit of the proposed method is first reflected in suppressing position drift, 
 while the azimuth solution is mainly improved indirectly through state coupling in the filtering framework. Taken together, these results show that the advantage
  of the proposed method does not lie in performing one-time geometric correction of the current position, but in constructing the local trajectory-to-route 
  matching result as a pseudo-position observation that can be incorporated into the EKF update, thereby enabling more stable road-level localization in different
   tunnel scenarios.

As also indicated by the Pos.Max values in Table~\ref{tab:scenario_comparison}, the proposed method achieves lower maximum position
 deviation than the baseline in all three scenarios. In particular, the reduction is most significant in the long-tunnel scenario,
  which further demonstrates the effectiveness of the proposed route-constrained correction under long-duration GNSS interruption.

\section{Conclusions}

To address  the accumulated localization drift for UGVs under severe GNSS degradation,
 this paper proposes a robust route-constrained state estimation method based on trajectory-to-route matching.
 When GNSS is unavailable, the route-referenced position is estimated based on the correspondence between the 
  historical dead reckoning trajectory and a local segment of the mission route, and is then incorporated into
   the EKF update as a pseudo-position observation. By doing so, road-level route constraints are injected into
    a unified state estimation framework without requiring additional sensors. Furthermore, route offset 
    compensation, trigger control, matching-quality validation, and single-step correction limiting strategies are 
    introduced to improve the engineering applicability of the method.

Experimental results demonstrate  that the proposed method can effectively suppress accumulated drift, mitigate the 
risk of significant localization deviations, and improve localization continuity as well as road-level usability 
in typical tunnel scenarios. The key advantage of the proposed method does not lie in performing one-time
 geometric correction of the current position, but in transforming the local trajectory-to-route matching 
 result into a pseudo-position that participates in the filtering update, thereby enabling more 
 consistent and stable degraded-navigation correction. Future work will further validate the proposed method using independent ground truth and investigate robust matching under complex road topologies, map-route bias, and adaptive pseudo-position confidence modeling.

\addtolength{\textheight}{-12cm}   






\end{document}